\documentclass[letterpaper, 10 pt, conference]{ieeeconf}  

\IEEEoverridecommandlockouts                              

\usepackage{graphics} 
\usepackage{float}
\usepackage{newfloat}
\usepackage[skins]{tcolorbox}
\usepackage{booktabs}
\usepackage{multirow}
\usepackage{tabularx}
\usepackage{dashrule}
\usepackage{todonotes}
\usepackage{url}
\usepackage{colortbl} 
\usepackage{array}    
\usepackage{afterpage}

\definecolor{yellowish}{RGB}{214,182,86}
\definecolor{greenish}{RGB}{130,179,102}

\newfloat{promptfloat}{!t}{pop}
\newtcolorbox[auto counter]{prompt}[2]{
    coltitle=black,
    label={prompt:#1},
    colback=gray!10,
    colframe=gray!20!gray,
    fonttitle=\bfseries,
    title=Prompt \thetcbcounter: #2,
    enhanced,
    fonttitle=\scriptsize, fontupper=\scriptsize, fontlower=\scriptsize, left=1mm, 
    right=1mm, 
    top=1mm, 
    bottom=1mm, 
    middle=1mm,
    arc=5pt,
    boxrule=0pt,
    borderline={0pt}{2pt}{solid}, 
    minipage boxed title*=-1.95em,
    attach boxed title to bottom center={yshift=2pt, yshift=0pt},
    boxed title style={enhanced, colback=white!55!white,
    boxrule=0pt, frame hidden}
}

\title{\LARGE \bf
Robots Can Multitask Too: Integrating a Memory Architecture and LLMs for Enhanced Cross-Task Robot Action Generation
}

\author{Hassan Ali$^{1}$, Philipp Allgeuer$^{1}$, Carlo Mazzola$^{2}$, Giulia Belgiovine$^{2}$, Burak Can Kaplan$^{1}$,\\ Lukáš Gajdošech$^{3}$ and Stefan Wermter$^{1}$
\thanks{Work supported by Horizon Europe project TERAIS (Grant number 101079338), the DFG Crossmodal Learning (TRR-169) project, and Study Abroad Graduate Scholarship by Ministry of National Education of Türkiye.}%
\thanks{$^{1}$Hassan Ali, Philipp Allgeuer, Burak Can Kaplan and Stefan Wermter are with the Knowledge Technology Group, Department of Informatics, University of Hamburg, Germany. Emails: {\tt\small \{hassan.ali, philipp.allgeuer, burak.can.kaplan, stefan.wermter\}@uni-hamburg.de}}%
\thanks{$^{2}$Carlo Mazzola and Giulia Belgiovine are with the CONTACT Unit, Italian Institute of Technology, Genoa, Italy. 
}%
\thanks{$^{3}$Lukáš Gajdošech is with the Faculty of Mathematics, Physics and Informatics, Comenius University, Slovakia. 
}%
}

\begin{document}

\maketitle
\thispagestyle{empty}
\pagestyle{empty}

\begin{abstract}
Large Language Models (LLMs) have been recently used in robot applications for grounding LLM common-sense reasoning with the robot's perception and physical abilities. In humanoid robots, memory also plays a critical role in fostering real-world embodiment and facilitating long-term interactive capabilities, especially in multi-task setups where the robot must remember previous task states, environment states, and executed actions. In this paper, we address incorporating memory processes with LLMs for generating cross-task robot actions, while the robot effectively switches between tasks. Our proposed dual-layered architecture features two LLMs, utilizing their complementary skills of reasoning and following instructions, combined with a memory model inspired by human cognition. Our results show a significant improvement in performance over a baseline of five robotic tasks, demonstrating the potential of integrating memory with LLMs for combining the robot's action and perception for adaptive task execution.
\end{abstract}

\section{INTRODUCTION}

Despite the physical limitations due to their embodiment, humanoid robots are particularly effective tools because of their anthropomorphic shape, which can significantly improve the versatility and effectiveness of robots, especially in environments designed for human interaction~\cite{damiano_2018_anthropomorphism}. 
Moreover, the humanoid physical shape of the robot supports collaboration with humans by making its actions more legible and predictable, which positively impacts safety and trust during interactions 
\cite{Vianello2021}. Inspired by the effectiveness and adaptability of human cognitive abilities, cognitive robotics aims to draw from more than embodiment when designing robotic platforms. Key attributes like human perception, selective attention, different types of memory, and other facets of human cognition are essential to improve the autonomy, flexibility, and capabilities of robots~\cite{Vernon2016}.


The development and utilization of foundational models in robotic platforms have been fundamental in driving advancement toward building autonomous artificial agents. Large Language Models (LLMs) have been used as symbolic reasoning instruments in various robotic applications 
such as reasoning about user's intentions in human-robot collaboration~\cite{ali2024_applesoranges}, generating robot action plans in grounded environments~\cite{singh2023_progpromt}, and task planning in environment-sensitive robot agents~\cite{sun2024_details}. 
Nevertheless, LLM reasoning alone is not yet sufficient for implementing the cognitive system of embodied artificial agents, capable of solving complex tasks and interacting with humans. 
Therefore, integrating LLMs with other cognitive abilities is essential for achieving this aim. Memory, in particular, is a capability that has been proven to contribute to the effectiveness of information processing in LLMs (for a survey, see \cite{zhang2024surveymemorymechanismlarge}). 

Taking inspiration from human cognition, several memory-related processes and mechanisms can be identified based on their specific roles and functionality, which can be implemented accordingly. 
In humans, although the precise taxonomy remains debated, short-term and long-term memory distinctively vary in the duration and capacity to retain information \cite{Cowan2008}. 
While declarative memory stores events and information, procedural memory has been conceptualized as the memory storing conditional instructions detailing actions based on circumstances \cite{Oberauer2010}. 
On the other hand, working memory integrates both representations from declarative and procedural memory \cite{Oberauer2010}, enabling the retention of essential information for an ongoing cognitive task, a process that is deeply connected with selective attention mechanisms \cite{Oberauer2019}. 

\begin{figure}[!t]
\parbox{\linewidth}{\centering\includegraphics[width=1.0\linewidth]{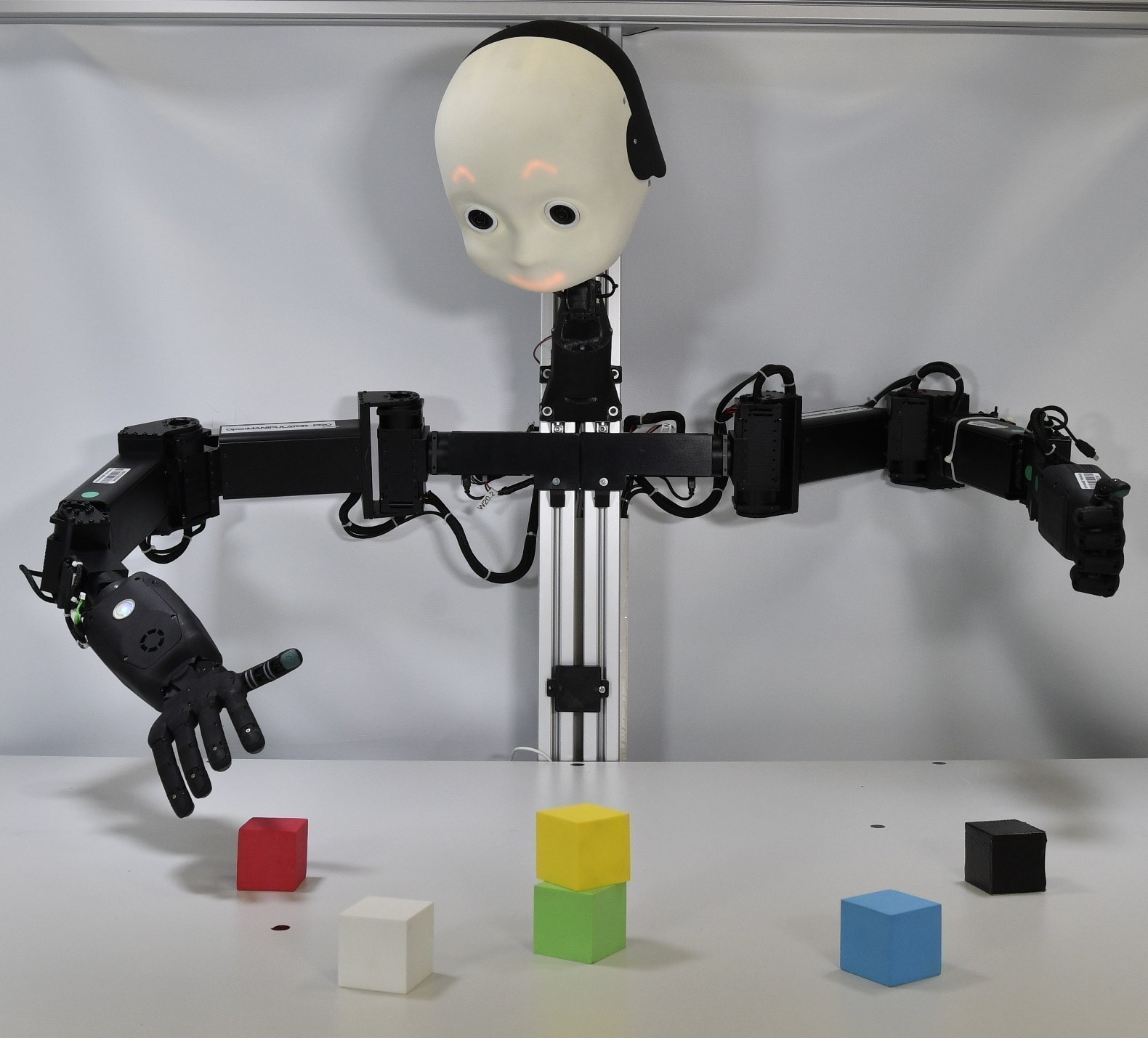}}
\caption{Our setup with the semi-humanoid NICOL robot.}
\label{fig:nicol_teaser}
\vspace{-5ex}
\end{figure}


Starting from this classification, in this work, we develop a hierarchical modular architecture to support the semi-humanoid NICOL collaborative robot (see Fig.~\ref{fig:nicol_teaser}) in reasoning with different kinds of memory functions. Thus, we enable the robot to solve the requests of a human agent by connecting 1) sensory information from the environment (perception), 2) LLM agent prompting (reasoning), and 3) task-appropriate robot actions produced as output. Although simply linking these three pillars might be sufficient for simple distinct tasks, the effectiveness might not be as high for more complex situations, like the multi-tasking required for natural interaction with humans. Therefore, we evaluate the impact of memory on multi-task execution and test our architecture with two different families of LLMs (OpenAI's GPT-3.5 and Meta's open-weight Llama~3). Our framework aligns and combines the environment data with task-specific memory cues, effectively bridging the gap between the robot's perception and action and laying solid ground for efficient task execution in contextually based interactions.

\section{Related Work}

Integrating sensory input with memory is essential in artificial systems as it enables them to retain sensory information for environment recognition and adaptive responses~\cite{wan2020_sensorymemory}. Moreover, memory plays a vital role in humanoid robots since it brings together 
conceptual knowledge gained through experience with the robot's behavior, enabling adjustment to complex interactive scenarios and tasks~\cite{peller-konrad2023_armarx}. Memory-based architectures have various applications such as transferring robot manipulation skills across different human environments~\cite{pohl2024_makeable}, memory-powered incremental learning with human-in-the-loop corrections~\cite{baermann2024_incrementallearning}, and supporting tasks like action-planning for localizing a target and navigating towards it~\cite{persiani2018_workingmemorymodel}. Existing work incorporates models of memory into robot tasks like planning with occluded objects~\cite{huang_2024_outofsight}, reasoning about object permanence~\cite{balkenius_2018_consciousrobot}, and learning human personal attributes like voices and faces~\cite{prescott_2019_timetravel}. However, such approaches model memory representations and reasoning as distinct components or separate processes while embedding memory directly within a reasoning backbone, like an LLM, 
is beneficial for context-aware decision-making.

LLMs have recently been used to support robots' reasoning and decision-making. However, relying solely on LLM's own internal memory is insufficient for handling complex multi-task reasoning demands of dynamic robotic environments. Some approaches address the limitation in LLM memory by injecting curated memory cues as contextual information into the LLM's prompts~\cite{zhang_2024_leaveit}, using previous structurally stored dialogues to enrich current LLM input~\cite{Zhong_2024_memorybank}, or caching previous conversations in a long-term memory storage~\cite{wang_2024_llmlongtermmemory}. Other methods rely on retraining or fine-tuning LLMs with contextual memory information~\cite{zhong_2022_trainingwithmemory}\cite{modarressi_2024_memllm}. However, relying on LLM context memory is constrained by context length and can lead to reasoning instability, while updating LLM parametric memory requires retraining and fine-tuning, causing cost and robustness concerns~\cite{zhang2024surveymemorymechanismlarge}. The limitations of LLM's own memory are also underscored in our experiments, highlighting the need for combining LLMs with efficient and up-to-date robust memory systems.

Our work addresses these limitations by incorporating LLMs for reasoning and action generation in a hierarchical framework. The idea of leveraging a hierarchy of LLMs to support reasoning and task specialization exists in the literature. For example, some approaches use multi-level LLMs to mitigate the latency issues of OpenAI APIs for real-time performance~\cite{liu_2024_overcooked}, combine LLM vision and textual processing capabilities~\cite{luu_2024_contextawarellmbasedsafecontrol}, and support efficiency in chatbot applications with levels of user-centered queries~\cite{nandkumar_2024_multilevelllm}. However, our LLM hierarchy differs from these approaches by utilizing a coordinator/worker LLM architecture, each with a different scope of the interaction context. Our structure also combines general LLM reasoning with specialized-task memory action execution, supporting multi-task reasoning and cognitive load reduction on each individual model.




\begin{figure}[!t]
\vspace{-1ex}
\parbox{\linewidth}
{\centering\includegraphics[width=1.0\linewidth]{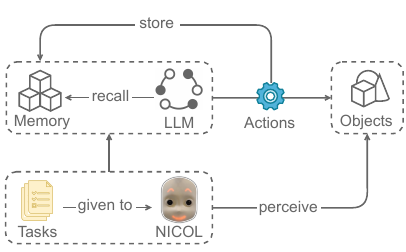}}
\caption{A simplified overview of our system's workflow. The workflow starts when the NICOL robot receives a given task. \textcolor{black}{The robot's sensory and task inputs are then fed to our proposed LLM-powered architecture.}}
\label{fig:workflow}
\vspace{-3ex}
\end{figure}

\section{Methodology}

We take inspiration from human cognition to integrate a memory framework with the robot's reasoning and action generation. In Fig.~\ref{fig:workflow}, we show a simplified overview of our system's workflow. The robot continuously observes the objects in its proximate environment through its visual perception and has specifications for 
pre-defined 
tasks. Upon 
a new task command, 
the system triggers the LLM to store and recall relevant memory components. 
Consequently, a suitable robot action 
is generated, applicable to the objects, and suitable to the task. Every time an 
action is performed, the memory records are updated, creating and maintaining a live memory repository and forming a continuous 
and adaptive feedback loop between the robot's actions, its memory, and the environment. We utilize two LLMs for 
managing the memory models, each optimized with prompts designed to achieve specific stages of the memory pipeline. 
In this section, we introduce our robot, its visual perception module, and present our proposed architecture and memory models.


\subsection{Robotic Platform: NICOL}
Our system runs on the Neuro-Inspired COLlaborator (NICOL)~\cite{kerzel2023_nicol} built in the Knowledge Technology group at the University of Hamburg. NICOL is a tabletop robotic platform with a humanoid head and two 8 DoF robotic arms. The robot features social cues like facial expressions as well as object and arm manipulation capabilities like grasping and neuro-genetic inverse kinematics~\cite{habekost2023_cycleik}. The NICOL platform is based on the Robot Operating System (ROS), facilitating integration with various sensory modules and deep learning models. It also supports open communication and collaboration with human users through an LLM-powered modular architecture~\cite{allgeuer2024_chatty} that combines sensory input like visually grounded open-vocabulary object detection (ViLD~\cite{gu2022_vild}) and speech recognition (Whisper~\cite{radford2022_whisper}) with actions like arm and head manipulation. Thus, it equips the robot with social and cognitive skills via its grounded LLM. 
In our approach, we extend NICOL's skills, but unlike previous work, we focus on 
adaptive action generation 
in robot multi-task scenarios. 

\begin{figure}[!t]
\vspace{1ex}
\parbox{\linewidth}{\centering\includegraphics[width=1.0\linewidth]{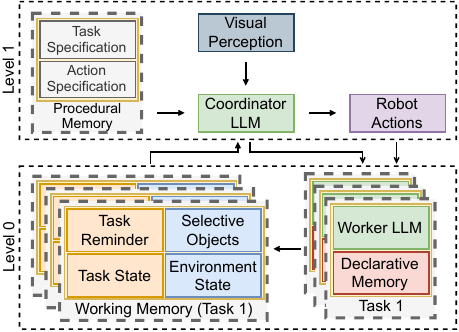}}
\caption{Our proposed system architecture consisting of two layers: \textit{level 0} and \textit{level 1}, utilizing the \textit{worker} and \textit{coordinator} LLM, respectively.}
\label{fig:architecture}
\vspace{-3ex}
\end{figure}

\subsection{NICOL's Visual Perception and Action Parsing}
We use NICOL's 4K camera located inside the robot's left eye to capture the environment visually and the objects on the table. We use the ViLD~\cite{gu2022_vild} object detector, enabling zero-shot object detection and recognition without pre-training. The NICOL's ViLD object detector has been optimized to work in real-time on continuous camera streams with high efficiency and reliability~\cite{allgeuer2024_chatty}. The detector generates bounding boxes for the objects on the table and their locations. All objects detected outside the area of interest, i.e., the robot's table, are removed. The objects detected are expressed by text so all object labels can be streamlined within the system's pipeline, making it compatible with an LLM application. NICOL can also parse textual LLM-generated actions and translate them into robot physical actions~\cite{allgeuer2024_chatty}. The robot can perform arm manipulations like pointing to objects on the table, grasping objects, and pushing objects to the user. For example, \textit{``point to the banana''} can be parsed and translated into corresponding physical actions performed by the robot's arm, i.e., an extended arm and extended index finger gesture to indicate the location of the banana on the table. The robot also has other head manipulation and facial expression capabilities. However, our system utilizes only the robot's arm manipulation skills in its action and task specifications.

\subsection{Proposed Architecture}
Our architectural design comprises a stack of two layers (cf. Fig.~\ref{fig:architecture}), which work collaboratively to achieve 
adaptive task execution. Each layer features an LLM corresponding to a specialized level of reasoning and memory capabilities but works in tandem with the other LLM to generate a set of robotic actions and achieve task-oriented goals. 
In the lower level (level 0), a \textit{worker} LLM is 
dedicated to creating 
and maintaining two memory variants (\textit{working} and 
\textit{declarative} memory) based on real-time inputs from the environment and interaction history, and 
supplying foundational information for the subsequent decision-making processes within the system. The worker LLM represents our instruction-based LLM which receives 
structured commands triggered by the \textit{coordinator} LLM, resulting in structured and precise outputs. 

The coordinator LLM in level 1 of the stack represents our reasoning LLM backbone which triggers the different functions of the worker LLM, 
integrates its output with the task's context, while focusing on complex task reasoning and decision-making. The coordinator LLM integrates input from the environment like the objects on the table and task-specific memory cues received from the worker LLM to generate robot-actionable commands and instructions, suitable to the robotic task at hand. 
For example, the output of the coordinator LLM in a tower building task might be an action like \textit{``put the yellow cube on top of the cube tower''} that requires reasoning and understanding of the task at hand, objects on the table, and overall context of previous interactions (which objects should be stacked in the tower, remaining cubes on the table, cubes already stacked in the tower, etc.). 

\begin{promptfloat}
\vspace{0.5ex}
\begin{prompt}{task_specs}{Task specifications as given to the coordinator LLM in the base prompt.}
\textbf{Separating task:} Your task is to sort objects in different boxes based on their properties. Fruits should go to box 1 while kitchenware and containers should go to box 2. For this task, you can place objects in the corresponding boxes using appropriate actions from the list of functions, for example: $<$move\_to\_box\_1(banana)$>$.\\[1ex]
\textbf{Arrangement task:} Your task is to place only the fruits inside the bowl. Do not put other objects inside the bowl. Use the $<$place\_in\_bowl$>$ action function to put the fruits one by one inside the bowl, for example, $<$place\_in\_bowl(banana)$>$.\\[1ex]
\textbf{Pointing task:} Your task is to point at objects in the following specific order, point at each yellow object then only when you have pointed at all the yellow objects, point at each red object on the table. Objects can be fruits, containers, or any other sort of object. Do not point at any object that is neither red nor yellow. Do not point at any object that is not on the table. At each step, check the remaining objects on the table to choose an object that fits the criteria. For example, to point at the banana, you can use $<$point(banana)$>$.
Similarly, repeat for all objects.\\[1ex]
\textbf{Recipe task:} Your task is to give all the ingredients for making a jello recipe with banana topping. The required objects are a bowl, jello, and banana. Do not give any other objects besides the required ingredients. Use the $<$give$>$ action function to give the objects one by one. Generate an action function for each object, for example $<$give(banana)$>$.\\[1ex]
\textbf{Tower task:} your task is to build a tower using colored cubes. Use the appropriate action function to stack cube objects on top of each other one by one. The order of the colored cubes does not matter. Generate an action function for each cube object, for example, $<$put\_on\_tower(red cube)$>$. 
But only use colored cubes for building the tower. Never use other objects for building the tower.\\[1ex]
When asked to perform this task, generate 
a set of actions to achieve it using the current objects. 
You should generate one action for each individual object.

\end{prompt}
\vspace{-5ex}
\end{promptfloat}

Collectively, the two layers (level 0 and level 1) enable the system to adapt to the changing environment as the objects on the table and task requirements are continuously changing, i.e., smoothly and seamlessly switching between tasks. Since both LLMs use text for input and output, the communication of information across the two layers
can be easily standardized and unified, thus, providing a robust framework for task execution that is 
responsive to the environment and also adaptive in terms of decision-making.

\begin{promptfloat}
\vspace{1ex}
\begin{prompt}{action_specs}{Action specifications as given to the coordinator LLM.}
        
        \textbf{$<$point(object)$>$:} Given a string of an object name, use your arms to point to that object on the table.\\[1ex]
        \textbf{$<$give(object)$>$:} Given a string of an object name, use your arms to push that object on the table and give it to the user. You can hand objects to the user with this function.\\[1ex]
        \textbf{$<$move\_to\_box\_1(object)$>$:} Given a string of an object name, use your arms to grasp that object on the table, lift it and drop it inside box 1 on the table. You can move objects to box 1 with this function.\\[1ex]
        \textbf{$<$move\_to\_box\_2(object)$>$:} Given a string of an object name, use your arms to grasp that object on the table, lift it and drop it inside box 2 on the table. You can move objects to box 2 with this function.\\[1ex]
        \textbf{$<$put\_on\_tower(object)$>$:} Given a string of an object name, use your arms to grasp that object on the table, lift it, and drop it on top of the cube tower. You can build a tower of cubes using this function.\\[1ex]
        \textbf{$<$place\_in\_bowl(object)$>$:} Given a string of an object name, use your arms to grasp that object on the table, lift it, and drop it inside the bowl on the table. You can place objects in the bowl with this function.
\end{prompt}
\vspace{-3ex}
\end{promptfloat}

\begin{promptfloat}
\begin{prompt}{memory_action_specs}{Memory functions for triggering the worker LLM.}
        \textbf{$<$retrieve\_working\_memory(task)$>$:} Given a string of a task name, use the following action format $<$retrieve\_working\_memory(task\_name)$>$ to retrieve the working memory. For instance, if you want to retrieve the working memory for separating, use $<$retrieve\_working\_memory(separating)$>$. Ensure that the output follows this format exactly.\\[1ex]
        \textbf{$<$retrieve\_declarative\_memory(task)$>$:} Given a string of a task name, use the following action format $<$retrieve\_declarative\_memory(task\_name)$>$ to retrieve the declarative memory. For instance, if you want to retrieve the declarative memory for separating, use $<$retrieve\_declarative\_memory(separating)$>$. Ensure that the output follows this format exactly.
\end{prompt}
\vspace{-5ex}
\end{promptfloat}

\subsection{LLMs for Instructions and Reasoning
}


We distinguish between the coordinator and worker LLMs optimized for reasoning and following instructions, respectively. 
The role of the coordinator LLM is to orchestrate the task execution process. At the system start, 
it receives a base prompt with specifications that ground the robot's knowledge of the possible tasks and their respective actions. This information represents the system's \textit{procedural memory}, informing the agent about the steps needed to achieve each task goal (see Prompt~\ref{prompt:task_specs} and Prompt~\ref{prompt:action_specs}). These specifications are modular and can be easily swapped or extended. 
The coordinator LLM receives the 
objects on the table from the open-vocabulary object detector 
and can then call two unique actions to invoke the worker LLM (see Prompt~\ref{prompt:memory_action_specs}) for 
triggering the rest of the memory process. 
After retrieving the 
output from the worker LLM, the knowledge can be injected within the existing 
specifications in order to prompt the coordinator LLM to generate a suitable robot action. 

At 
level 0, the worker LLM is responsible for parsing and interpreting 
the environment input (passed from the coordinator LLM), and thus, dynamically managing the real-time sensory perception of the robot. 
Upon receiving the instruction from the coordinator LLM, it creates the working memory, which represents the selective 
attention of the robot. 
Additionally, it maintains the declarative memory, which represents facts about the history of interaction over a time period. Unlike the coordinator LLM which has an overview of all the interactions, i.e., its conversations include all the previous dialogues, the worker LLM receives only a single prompt at a time without any previous dialogue or context.




\begin{promptfloat}
\vspace{1ex}
\begin{prompt}{working_memory_example}{An example prompt to extract the working memory (separating task).}
\textbf{Prompt Input:}\\
Separating Task: the task is to move fruits to box 1, containers and kitchenware objects to box 2. $\longrightarrow$ \textit{Task Description} \\
Name the objects that are relevant to the given task from the following: $\longrightarrow$ \textit{LLM Instruction}\\
apple, banana, cup, bowl, baseball, pear $\longrightarrow$ \textit{Visual Input}\\
Output a list of object names separated by a comma and without any extra text. If order is important to the task then output the object names in the correct order. $\longrightarrow$\textit{Output Format}\\[1ex]
\textbf{Prompt Output:}\\
apple, banana, cup, bowl, pear
\end{prompt}
\vspace{-3ex}
\end{promptfloat}



\begin{promptfloat}
\begin{prompt}{longterm_memory_example}{An example prompt to extract the declarative memory (separating task).}
\textbf{Prompt Input:}\\
Log Entry:\\
Action: $<$move\_to\_box\_1(pear)$>$\\
Box 1: 1. pear \\
Remaining Objects: 1. apple 2. banana 3. cup 4. bowl 5. baseball\\
Log Entry:\\
Action: $<$move\_to\_box\_1(apple)$>$\\
Box 1: 1. pear 2. apple \\
Remaining Objects: 1. banana 2. cup 3. bowl 4. baseball\\
Log Entry:\\
Action: $<$move\_to\_box\_2(bowl)$>$\\
Box 2: 1. bowl \\
Remaining Objects: 1. banana 2. cup 3. baseball\\
$\longrightarrow$\textit{Log File}\\
Given the sequence of log entries, extract the final list of objects in box 1 from the last log entry. $\longrightarrow$\textit{LLM Instruction}\\ Output a list of object names separated by a comma and without any extra text. $\longrightarrow$\textit{Output Format}\\[1ex]
\textbf{Prompt Output:}\\
pear, apple
\end{prompt}
\vspace{-5ex}
\end{promptfloat}

\subsection{Working Memory}



The worker LLM extracts 
and refines 
the working memory, which is a compact representation of crucial information to task completion. Each working memory is specialized for a specific task 
without knowledge of other tasks, and consists 
of several task- and environment-related cues. Task-related cues are: 1) \textit{Task Reminders}: additional short descriptions of the task actions and desired goal, and 2) \textit{Task State}: which 
tracks the current state of the objects in relevance to the current task progress. 
For example, in a fruit arrangement task (the fruits must be placed in the bowl), it might include information such as \textit{``the banana and apple are in the bowl''}.

The working memory also includes a \textit{selective object} state extracted by combining the information of the declarative and procedural memory. Specifically, it receives all the objects on the table 
and outputs only the task-relevant ones
, representing the robot's visual selective attention. 
In Prompt~\ref{prompt:working_memory_example}, we highlight an example of the separating task (fruits and dishware items must be placed in different bins). The first part of the prompt is the task description, which is then combined with the visual input 
identifying the objects: 
apple, banana, cup, bowl, baseball, pear. The prompt's final part sets rules for the worker LLM on output formulation, ensuring the working memory remains consistent before passing it back to the coordinator LLM. As a result, only 
necessary objects are retained, i.e., objects not task-relevant are removed from the working memory (the baseball in the given example). 



\subsection{Declarative Memory} 



When the robot modifies the environment with object manipulation actions, these changes are logged into the system's declarative memory, ensuring that the robot's internal state through its memory is precisely synchronous with the external world. Since all the robot actions are recorded, the system always has knowledge of the interaction history over any duration of time. 
We store this memory 
persistently as log files, one file per task. Similar to working memory, declarative memory is specialized to one task and does not contain information about other tasks. As a task newly 
starts, the interaction history is empty and 
data is accumulated as the task progresses. \textcolor{black}{Although the memory is extended dynamically, since the robot generates one action per object, the memory size (number of entries) can be at most equal to the number of objects}. If the task gets interrupted, i.e., the robot is asked to switch to another one, the interaction history 
is kept safe. 
When the robot 
continues the interrupted task, it 
retrieves essential information like 
previous actions, environment state, and task state at the interruption point.


The worker LLM is responsible for creating and maintaining the logs of the declarative memory. 
An example using the separating task is shown in Prompt~\ref{prompt:longterm_memory_example} (for simplicity, we only show the information extraction of box 1). The worker LLM is given the persistent task logs and prompted to extract the remaining objects on the table (environment state) and the objects in box 1 and box 2 (task state). Similar to the working memory, the prompt has a clearly defined output structure for consistency of the information in the system 
across the two LLMs. The log file is structured by entries ordered by the executed actions, which 
allows the LLM to efficiently 
identify the most recent environment and task states. 

\begin{table}[!t]
\vspace{1ex}
\caption{Summary of the required reasoning and actions per task
}
\begin{tabular}{@{}llll@{}}
\toprule
\textbf{Task}      & \textbf{Input Objects}                                                                                                & \textbf{Reasoning}                                                                                                            & \textbf{Action (s)}                                                          \\ \midrule
\textbf{Separate}   & \begin{tabular}[c]{@{}l@{}}apple, banana, cup, \\ bowl, baseball, pear\end{tabular}       

& \begin{tabular}[c]{@{}l@{}}fruits $\rightarrow$ \textit{box 1}, \\ kitchenware$\rightarrow$\textit{box 2}\\others $\rightarrow$ \textit{table}\end{tabular}
& \begin{tabular}[c]{@{}l@{}}move\_to\_box\_1\\ move\_to\_box\_2\end{tabular} \\[2ex]
\hline\addlinespace[0.8ex]
\textbf{Arrange} & \begin{tabular}[c]{@{}l@{}}apple, banana, can, \\ lemon, orange, pear\end{tabular}     

& \begin{tabular}[c]{@{}l@{}}fruits $\rightarrow$ \textit{bowl}\\ others $\rightarrow$ \textit{table}\end{tabular}
                                                            & place\_in\_bowl                                                              \\[1.5ex]
                                                            \hline\addlinespace[0.8ex]

\textbf{Point}   & \begin{tabular}[c]{@{}l@{}}\textcolor{red}{apple}, \textcolor{red}{can}\\ \textcolor{yellowish}{lemon}, \textcolor{yellowish}{banana}\\ \textcolor{orange}{orange}, \textcolor{greenish}{pear}\end{tabular}                                                                             & \begin{tabular}[c]{@{}l@{}l@{}}order of pointing:\\yellow $\rightarrow$ red\\ others $\rightarrow$ \textit{table}\end{tabular} & point \\[3ex]
\hline\addlinespace[0.8ex]                                                  
\textbf{Recipe}    & \begin{tabular}[c]{@{}l@{}}apple, banana, can \\ bowl, jello, pear\end{tabular}                                                                                 & \begin{tabular}[c]{@{}l@{}}recipe objects: \\ bowl, jello, banana\end{tabular}                                                                                          & give                                                                         \\[2ex]\hline\addlinespace[0.8ex]

\textbf{Tower}   & \begin{tabular}[c]{@{}l@{}}\textcolor{red}{cube 1}, \textcolor{yellowish}{cube 2}\\ \textcolor{blue}{cube 3}, cube 4 (b)\\ cube 5 (w), \textcolor{greenish}{cube 6}\end{tabular}                                                                            & \begin{tabular}[c]{@{}l@{}}colored $\rightarrow$ Tower\\ black/white$\rightarrow$\textit{table}\end{tabular} & put\_on\_tower                                                             \\ \bottomrule
\end{tabular}
\label{table:tasks_summary}
\vspace{-3ex}
\end{table}

\section{Experiments and Evaluation}
We evaluate our system using five robotic tasks, each requiring a set of objects as input and an appropriate sequence of robot actions as output 
to achieve the task expected goal. Each task involves both a cognitive aspect, where the robot must analyze the objects 
and reason about the task definition to identify the 
ones required to achieve it, and a physical aspect, in which the robot performs the appropriate object manipulation function. 
The objects necessary to perform tasks vary per task, however, we limit the total input to six objects since our early experiments showed that an input size of six objects is most reasonable given the limited number of tokens the LLM can process within its context window. We use YCB objects for the tasks in addition to colored cubes for the tower task.
An overview of the tasks is as follows: 1) \textit{Separating Task:} the robot places fruits and dishwasher items in different containers, 2) \textit{Arrangement Task:} the fruits must be arranged by placing them inside a bowl, 3) \textit{Pointing Task:} the robot points to all the yellow objects on the table and then to all the red objects, 4) \textit{Recipe Task:} objects for making a jello recipe with banana topping are given to the user, and 5) \textit{Tower Task:} the robot builds a tower by stacking only colored cubes on top of each other. A summary of the tasks is provided in Table~\ref{table:tasks_summary}. Our system runs on the NICOL robot, but for evaluation, we disable the robot's physical control since we focus on evaluating our system's action-generating aspect and not the NICOL platform components. 

\begin{figure}[!t]
\vspace{1ex}
\centering
\parbox{\linewidth}{\centering\includegraphics[width=1.0\linewidth]{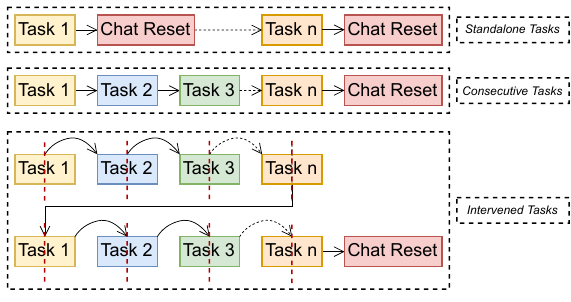}}
\caption{Our model evaluation scheme has three different task execution modes, each resetting the LLM chat at a different interaction point.}
\label{fig:evaluation}
\vspace{-3ex}
\end{figure}

We conducted a total of 50 trials for each experiment using two LLMs: ChatGPT-3.5-Turbo-0125 and 
Llama~3. The GPT-3.5 model has 175B parameters and a context window length of approximately 
16k tokens. We opted for GPT-3.5 over the GPT-4.0 models since it is more efficient in terms of cost and speed. Experiments with this model were conducted using the API services of OpenAI. In contrast, Llama~3 has 70B parameters and a context window size of 
8k tokens. An advantage of open-weight models is that they allow potentially controlling the seed values, supporting result reproducibility. All the Llama experiments were conducted using a single NVIDIA A100 GPU with 80 GB of VRAM. Due to the GPU's capacity constraints, we utilized the 8-bit quantized version of Llama~3, requiring approximately 72 GB of VRAM. This model is provided by Ollama\footnote{\url{https://ollama.com/}} and compatible with Langchain\footnote{\url{https://python.langchain.com/v0.2/docs/integrations/llms/ollama/}}.
We used identical parameters for 
both LLMs to ensure a fair comparison. The $temperature$ was set to 0.2 for controlled results. Given that consistency is already regulated by the temperature value, we set $top\_p$ to 1 to avoid eliminating any potential candidate responses. Both the $frequency\_penalty$ and $presence\_penalty$ were set to 0, acknowledging that the same tokens may recur due to the defined scope of objects and actions in the experiment. 

\subsection{Single and Multi-task Scenarios}
Our evaluation scheme consists of three modes (as sketched in Fig.~\ref{fig:evaluation}). \textit{Standalone} mode refers to running each task independently, followed by a complete LLM chat history reset. This experiment represents the baseline performance for each of the defined tasks. In \textit{consecutive} mode, we run the tasks back to back in a sequence and reset the chat only after executing the last task. In the \textit{intervened} mode, the system performs a multi-task scenario, in which each task is interrupted at an intervention point (defined at a 50\% task completion based on the total number of required actions). After each intervention point, the system switches to the next task and back to Task 1 after the completion of the last task. A chat reset follows a 100\% completion of the last task. The consecutive and intervened scenarios require memory components, and thus, we evaluate them with and without memory for comparison. We present the results of our experiments in the next subsections, using the following metrics: 1) \textit{Success Rate}: the ratio of the correctly completed tasks to the total trials. For a task to be considered complete, the system must generate all the required actions correctly, 2) \textit{Task Retention}: assesses whether the system can correctly identify and recall the state of the task after its completion, e.g., which objects are in box 1 and box 2 and, 3) \textit{Environment Retention}: verifies whether the system can correctly recognize the state of the environment after the task completion, i.e., which objects remain on the table.



\subsection{Results for Standalone Tasks}
This experiment is our single-task scenario, meaning the system is configured to execute one task at a time with knowledge limited to the task at hand and its actions. Therefore, the system does not require memory of previous interactions and runs with the memory components disabled. As shown in Table~\ref{table:results_1_standalone}, the system completes the tasks with a high success rate as they are presented individually, and with comparable performance across both LLMs. 
Furthermore, the system 
maintains task- and environment-related information with high accuracy when evaluated at the end of each task, even reaching a perfect score for some tasks. 
The Llama~3 success rate slightly dropped for the arrangement and recipe tasks. 
In the arrangement task, Llama~3 occasionally generated a batch of actions instead of a single action per prompt. Despite the soundness of the generated actions, generating multiple actions at a time violates our strictly defined evaluation scripts. In the recipe task, Llama~3 occasionally generated actions for giving the wrong toppings, likely due to the model's training containing real data of recipes with jello and fruits. However, this setup demonstrates that the tasks are inherently manageable for the LLM hierarchy to handle when presented in a standalone manner. Thus, this scenario represents our baseline of task performance 
compared to the more complex evaluation modes: consecutive and intervened.

\begin{table}[]
\vspace{1ex}
\centering
\caption{Results of the \textbf{standalone tasks} experiments
}
\begin{tabular}{@{}lcccc@{}}
\toprule
                                                     \textbf{Task}  & \textbf{Model} & \textbf{\begin{tabular}[c]{@{}c@{}}Success\\Rate\end{tabular}}     & \textbf{\begin{tabular}[c]{@{}c@{}}Task\\Retention\end{tabular}}& \textbf{\begin{tabular}[c]{@{}c@{}}Environment\\Retention\end{tabular}} \\ \midrule
\multirow{2}{1cm}{\textbf{Separate}}                        
                                                     & GPT~3.5 & 0.98                     & 0.99                & 1.00                                              \\
                                                     & Llama~3 & 0.98                     & 0.98                & 1.00                                              \\[0.5ex]
                                                     
                                                     \hline\addlinespace[0.8ex]
\multirow{2}{1cm}{\textbf{Arrange}}
                                                     & GPT~3.5 & 0.92                     & 0.94                 & 0.96                                             \\
                                                     & Llama~3 & 0.86                     & 0.98                & 0.96                                              \\[0.5ex]
                                                     
                                                     \hline\addlinespace[0.8ex]
\multirow{2}{1cm}{\textbf{Point}} 
                                                     & GPT~3.5 & 1.00                     & 1.00                 & 1.00                                              \\
                                                     & Llama~3 & 0.98                     & 0.98                & 0.98                                              \\[0.5ex]
                                                     \hline\addlinespace[0.8ex]
\multirow{2}{1cm}{\textbf{Recipe}}  
                                                     & GPT~3.5 & 0.96                     & 1.00                 & 1.00                                              \\
                                                     & Llama~3 & 0.86                     & 0.96                & 0.96                                              \\[0.5ex]
                                                     \hline\addlinespace[0.8ex]
\multirow{2}{1cm}{\textbf{Tower}}
                                                     & GPT~3.5 & 1.00                     & 1.00                 & 1.00                                              \\
                                                     & Llama~3 & 0.98                     & 0.98                & 1.00                                              \\
                                                     
                                                     \bottomrule
                                                     
\end{tabular}
\label{table:results_1_standalone}
\vspace{-3ex}
\end{table}

\subsection{Results for Consecutive Tasks}
In this setup, the system has knowledge of all the defined tasks and possible actions through its 
declarative memory. Hence, the LLM must not only reason about the task requirements but also choose the appropriate actions. In this experiment, we observe that running the tasks 
consequently presents a considerable challenge to the 
LLMs when tuned \textit{without the working memory} (cf. Table~\ref{table:results_2_consecuetive_tasks}). The system achieves average performance for some of the tasks (separating and recipe), however, the success rate and retention drop significantly for most tasks compared to the standalone experiment. 
The pointing task 
is particularly challenging since the LLM must reason about the order of objects, 
not only their type, as well as the appropriate pointing action. Similarly, the tower task is challenging to the system, and we observe 
failures in correctly reasoning that black and white cubes (not colored) must not be stacked in the tower.

\begin{figure}[!b]
\parbox{\linewidth}{\centering\includegraphics[width=1.0\linewidth]{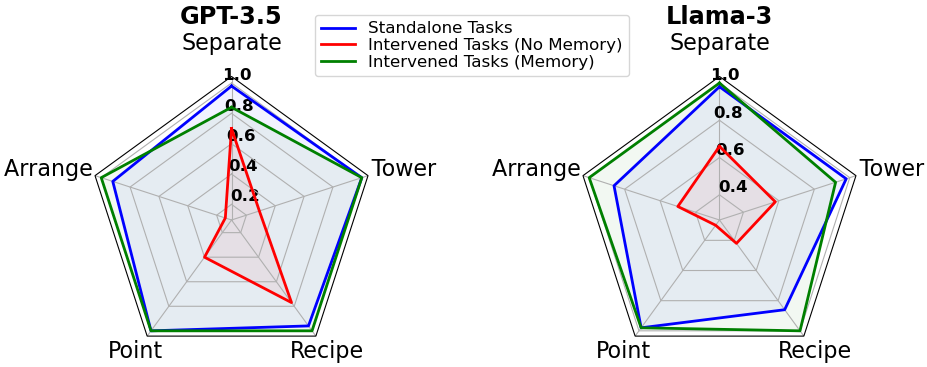}}
\caption{Success rate with both LLMs in the \textbf{consecutive setup} (with and without memory) in comparison to the baseline (standalone setup).}
\label{fig:spider_consecutive}
\end{figure}

In contrast, we report a noticeable boost in the success rate and retention metrics when running the experiment \textit{with the working memory enabled} (see Table~\ref{table:results_2_consecuetive_tasks}). Both LLM variants perform comparably and even reach a perfect score for some of the tasks, e.g., arrangement. Noticeably, Llama~3 outperforms ChatGPT~3.5 on the separating task due to errors in the coordinator bypassing the worker LLM output and generating redundant unnecessary actions, negatively influencing its overall success and retention values. Overall, both LLMs achieve success rates that approximate the baseline (standalone setup) and even surpass it for tasks like arrangement and recipe (see Fig.~\ref{fig:spider_consecutive}). This experiment shows a significant improvement in performance due to the utilization of the worker LLM and working memory, thus, suggesting that the coordinator LLM becomes a bottleneck in 
action generation over a longer period of interaction time.

\begin{table}[]
\vspace{1ex}
\centering
\caption{Results of the \textbf{consecutive tasks} experiment without and with memory (highlighted)}
\begin{tabular}{@{}lccccccc@{}}
\toprule
\textbf{Task}                     & \cellcolor{white!25}\textbf{Model}               & \multicolumn{2}{c}{\textbf{\begin{tabular}[c]{@{}c@{}}Success\\ Rate\end{tabular}}} & \multicolumn{2}{c}{\textbf{\begin{tabular}[c]{@{}c@{}}Task\\ Retention\end{tabular}}} & \multicolumn{2}{c}{\textbf{\begin{tabular}[c]{@{}c@{}}Environment\\ Retention\end{tabular}}} \\ \midrule
\multirow{2}{*}{\textbf{Separate}}    & \multicolumn{1}{l|}{GPT~3.5} & 0.70                           & \multicolumn{1}{l|}{\cellcolor{pink!25}0.84}                          & 0.61                            & \multicolumn{1}{l|}{\cellcolor{pink!25}0.92}                           & 0.22                                          & \multicolumn{1}{c}{\cellcolor{pink!25}0.92}                                         \\
                                  & \multicolumn{1}{l|}{Llama~3} & 0.66                           & \multicolumn{1}{l|}{\cellcolor{pink!25}1.00}                          & 0.78                            & \multicolumn{1}{l|}{\cellcolor{pink!25}1.00}                           & 0.20                                          & \multicolumn{1}{c}{\cellcolor{pink!25}1.00}                                         \\ \midrule
\multirow{2}{*}{\textbf{Arrange}} & \multicolumn{1}{l|}{GPT~3.5} & 0.14                           & \multicolumn{1}{l|}{\cellcolor{pink!25}1.00}                          & 0.14                            & \multicolumn{1}{l|}{\cellcolor{pink!25}1.00}                           & 0.72                                          & \multicolumn{1}{c}{\cellcolor{pink!25}1.00}                                         \\
                                  & \multicolumn{1}{l|}{Llama~3} & 0.50                           & \multicolumn{1}{l|}{\cellcolor{pink!25}1.00}                          & 0.50                            & \multicolumn{1}{l|}{\cellcolor{pink!25}1.00}                           & 0.50                                          & \multicolumn{1}{c}{\cellcolor{pink!25}1.00}                                         \\ \midrule
\multirow{2}{*}{\textbf{Point}}   & \multicolumn{1}{l|}{GPT~3.5} & 0.40                           & \multicolumn{1}{l|}{\cellcolor{pink!25}1.00}                          & 0.40                            & \multicolumn{1}{l|}{\cellcolor{pink!25}1.00}                           & 0.40                                          & \multicolumn{1}{c}{\cellcolor{pink!25}1.00}                                         \\
                                  & \multicolumn{1}{l|}{Llama~3} & 0.30                           & \multicolumn{1}{l|}{\cellcolor{pink!25}0.98}                          & 0.40                            & \multicolumn{1}{l|}{\cellcolor{pink!25}1.00}                           & 0.28                                          & \multicolumn{1}{c}{\cellcolor{pink!25}1.00}                                         \\ \midrule
\multirow{2}{*}{\textbf{Recipe}}  & \multicolumn{1}{l|}{GPT~3.5} & 0.77                           & \multicolumn{1}{l|}{\cellcolor{pink!25}1.00}                          & 0.68                            & \multicolumn{1}{l|}{\cellcolor{pink!25}1.00}                           & 0.68                                          & \multicolumn{1}{c}{\cellcolor{pink!25}1.00}                                         \\
                                  & \multicolumn{1}{l|}{Llama~3} & 0.42                           & \multicolumn{1}{l|}{\cellcolor{pink!25}1.00}                          & 0.82                            & \multicolumn{1}{l|}{\cellcolor{pink!25}1.00}                           & 0.18                                          & \multicolumn{1}{c}{\cellcolor{pink!25}1.00}                                         \\ \midrule
\multirow{2}{*}{\textbf{Tower}}   & \multicolumn{1}{l|}{GPT~3.5} & 0.29                           & \multicolumn{1}{l|}{\cellcolor{pink!25}1.00}                          & 0.28                            & \multicolumn{1}{l|}{\cellcolor{pink!25}1.00}                           & 0.32                                          & \multicolumn{1}{c}{\cellcolor{pink!25}1.00}                                         \\
                                  & \multicolumn{1}{l|}{Llama~3} & 0.58                           & \multicolumn{1}{l|}{\cellcolor{pink!25}0.92}                          & 0.68                            & \multicolumn{1}{l|}{\cellcolor{pink!25}0.98}                           & 0.36                                          & \multicolumn{1}{c}{\cellcolor{pink!25}0.94}                                         \\ \bottomrule
\end{tabular}
\label{table:results_2_consecuetive_tasks}
\vspace{-4ex}
\end{table}

\subsection{Results for Intervened Tasks}
We evaluate the declarative and working memory by creating intervention points, i.e., interruptions, in which the execution of a task is paused, and the system must switch to the next task. Each trial in this experiment involves two execution rounds. In the first round, only 50\% progress of each task is accomplished, and in the second, the system 
resumes every paused task from the interruption points until full completion. \textcolor{black}{The 50\% intervention is chosen for simplicity; however, the system behaves similarly at any interruption point.} For consistency, we run the tasks by their definition order: separating, arrangement, pointing, recipe, and tower. However, the task order does not influence the experiment since each task has its own input and runs independently, i.e., one task success or failure does not impact other tasks' performance. 
We run the intervention mode twice: without memory and with declarative and working memory. 
Similar to the consecutive mode, the tasks are challenging to 
both LLMs without memory (cf.~Table~\ref{table:results_3_intervened_tasks}). Noticeably, the recipe task performance drops for GPT~3.5 
due to errors in the coordinator LLM, generating 
noisy 
actions on wrong objects. 
Moreover, the performance of Llama~3 is below average for all tasks. The main 
error source is the LLM ``forgetting'' the task specifications due to the longer interaction duration, and thus, generating false actions or no actions. 
Also, the logical task reasoning was challenging, and the coordinator LLM generated incorrect actions, 
e.g., pointing to an object neither yellow nor red, placing a non-fruit object in the bowl, or giving the wrong object in the recipe task. 

\begin{figure}[!b]
\parbox{\linewidth}{\centering\includegraphics[width=1.0\linewidth]{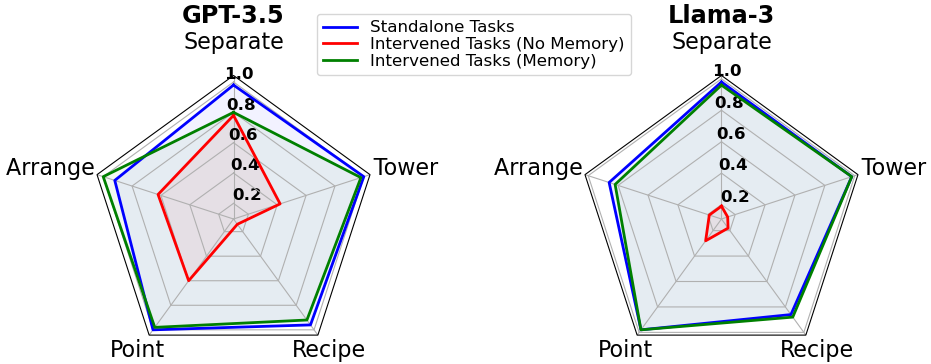}}
\caption{Success rate with both LLMs in the \textbf{intervened setup} (with and without memory) in comparison to the baseline (standalone setup).}
\label{fig:spider_intervened}
\end{figure}

Here again, our experiment demonstrates noticeable improvement in the success rate, task retention, and environment retention scores across both LLMs (see~Table~\ref{table:results_3_intervened_tasks}). The scores for the pointing, recipe, and tower tasks are comparable for both models and approximate the baseline (see Fig.~\ref{fig:spider_intervened}). Interestingly, Llama~3 scored higher success and retention values in the separating task due to the task grounding of GPT~3.5 leading to faulty commonsense reasoning, thus, occasionally generating wrong actions such as $<$move\_to\_box\_2(pear)$>$ instead of $<$move\_to\_box\_1(pear)$>$. We also observed Lama~3 generating a wrong list of objects in the working memory of the arrangement task, which explains the slight drop in its performance. However, when comparing against the previous experiment (consecutive with memory), this seems to be a challenge only when the system must provide this information in the second round of task execution. We hypothesize this to be due to the context size of Llama~3 being roughly half the size of this ChatGPT model (8k vs 16k) since this mode requires multiple rounds. We also observed high variability in the output with these answers, hinting that the 
parameters chosen for Llama~3 could be further optimized for more stable output (ex: we used $top\_p$ of 1 for both models).

\begin{table}[]
\vspace{1ex}
\centering
\caption{Results of the \textbf{intervened tasks} experiment without and with memory (highlighted)}
\begin{tabular}{@{}lccccccc@{}}
\toprule
\textbf{Task}                     & \cellcolor{white!25}\textbf{Model}               & \multicolumn{2}{c}{\textbf{\begin{tabular}[c]{@{}c@{}}Success\\ Rate\end{tabular}}} & \multicolumn{2}{c}{\textbf{\begin{tabular}[c]{@{}c@{}}Task\\ Retention\end{tabular}}} & \multicolumn{2}{c}{\textbf{\begin{tabular}[c]{@{}c@{}}Environment\\ Retention\end{tabular}}} \\ \midrule
\multirow{2}{*}{\textbf{Separate}}    & \multicolumn{1}{l|}{GPT~3.5} & 0.78                           & \multicolumn{1}{l|}{\cellcolor{pink!25}0.80}                          & 0.58                            & \multicolumn{1}{l|}{\cellcolor{pink!25}0.75}                           & 0.60                                          & \multicolumn{1}{c}{\cellcolor{pink!25}0.98}                                         \\
                                  & \multicolumn{1}{l|}{Llama~3} & 0.19                           & \multicolumn{1}{l|}{\cellcolor{pink!25}0.96}                          & 0.15                            & \multicolumn{1}{l|}{\cellcolor{pink!25}0.99}                           & 0.66                                          & \multicolumn{1}{c}{\cellcolor{pink!25}0.98}                                         \\ \midrule
\multirow{2}{*}{\textbf{Arrange}} & \multicolumn{1}{l|}{GPT~3.5} & 0.62                           & \multicolumn{1}{l|}{\cellcolor{pink!25}1.00}                          & 0.44                            & \multicolumn{1}{l|}{\cellcolor{pink!25}1.00}                           & 0.74                                          & \multicolumn{1}{c}{\cellcolor{pink!25}1.00}                                         \\
                                  & \multicolumn{1}{l|}{Llama~3} & 0.19                           & \multicolumn{1}{l|}{\cellcolor{pink!25}0.82}                          & 0.50                            & \multicolumn{1}{l|}{\cellcolor{pink!25}0.82}                           & 0.12                                          & \multicolumn{1}{c}{\cellcolor{pink!25}0.82}                                         \\ \midrule
\multirow{2}{*}{\textbf{Point}}   & \multicolumn{1}{l|}{GPT~3.5} & 0.60                           & \multicolumn{1}{l|}{\cellcolor{pink!25}0.98}                          & 0.70                            & \multicolumn{1}{l|}{\cellcolor{pink!25}0.98}                           & 0.72                                          & \multicolumn{1}{c}{\cellcolor{pink!25}0.98}                                         \\
                                  & \multicolumn{1}{l|}{Llama~3} & 0.28                           & \multicolumn{1}{l|}{\cellcolor{pink!25}0.98}                          & 0.24                            & \multicolumn{1}{l|}{\cellcolor{pink!25}0.98}                           & 0.16                                          & \multicolumn{1}{c}{\cellcolor{pink!25}0.98}                                         \\ \midrule
\multirow{2}{*}{\textbf{Recipe}}  & \multicolumn{1}{l|}{GPT~3.5} & 0.14                           & \multicolumn{1}{l|}{\cellcolor{pink!25}0.92}                          & 0.12                            & \multicolumn{1}{l|}{\cellcolor{pink!25}0.92}                           & 0.12                                          & \multicolumn{1}{c}{\cellcolor{pink!25}0.92}                                         \\
                                  & \multicolumn{1}{l|}{Llama~3} & 0.18                           & \multicolumn{1}{l|}{\cellcolor{pink!25}0.88}                          & 0.42                            & \multicolumn{1}{l|}{\cellcolor{pink!25}0.92}                           & 0.32                                          & \multicolumn{1}{c}{\cellcolor{pink!25}0.98}                                         \\ \midrule
\multirow{2}{*}{\textbf{Tower}}   & \multicolumn{1}{l|}{GPT~3.5} & 0.42                           & \multicolumn{1}{l|}{\cellcolor{pink!25}0.98}                          & 0.46                            & \multicolumn{1}{l|}{\cellcolor{pink!25}0.98}                           & 0.44                                          & \multicolumn{1}{c}{\cellcolor{pink!25}0.98}                                         \\
                                  & \multicolumn{1}{l|}{Llama~3} & 0.15                           & \multicolumn{1}{l|}{\cellcolor{pink!25}0.98}                          & 0.16                            & \multicolumn{1}{l|}{\cellcolor{pink!25}0.98}                           & 0.48                                          & \multicolumn{1}{c}{\cellcolor{pink!25}0.98}                                         \\ \bottomrule
\end{tabular}
\label{table:results_3_intervened_tasks}
\vspace{-4ex}
\end{table}

\section{Discussion and Conclusion}

In this work, we proposed an architecture for combining 
cognitively inspired memory processes with a hierarchical LLM-powered framework 
for enabling the robot to smoothly and 
effectively switch between tasks. 
Our layered approach leverages the efficiency and accuracy of LLMs at executing instructions and their more advanced 
reasoning skills, resulting in the ability to maintain contextually rich interactions across multiple tasks. Our system has been applied to the physical NICOL robot but generally has various applications for increasing the productivity of human-robot collaboration, e.g., enabling the robot to simultaneously perform multiple tasks 
and manage various high-level goals in parallel. Our system can be extended with more task and action specifications and can be used for adaptive task execution that considers the interaction context at runtime. Thanks to our modular architecture, the system can integrate different LLM types, e.g., a powerful LLM for reasoning and a smaller lower-cost LLM for executing instructions, supporting cost and time efficiency while maintaining effective task handling.

Our experiments showed 
a 
challenge for the current 
LLMs (ChatGPT and Llama) to manage different tasks and goals at the same time by relying on the LLM's own internal memory or the context of previous dialogues in the prompt. However, our implemented memory process (declarative, procedural, and working memory) 
demonstrated a significant boost in the performance of both models compared to our baseline. \textcolor{black}{Although LLMs enable the flexibility of task definition and reasoning without explicit programming, their fixed context size can restrict the number of tasks and objects that can be managed simultaneously. Also, our experiments showed that LLMs are still prone to errors like occasionally generating redundant or incorrect actions. However,} in future work, memory-based architectures like our proposed one can be expected to be beneficial for human-robot collaboration besides action generation. For example, we plan research on managing high-level goals in multi-party robotic scenarios, where humans are interacting with the robot, each having their own goal. As LLMs with larger context windows become more available and accessible, we will also evaluate our system with a larger number of tasks and input objects.

\bibliographystyle{IEEEtran}
\bibliography{IEEEabrv, main}

\end{document}